\documentclass{article}
\usepackage{times}
\usepackage{graphicx} % more modern
\usepackage{natbib}
\usepackage[nohyperref,accepted]{icml2017}
\usepackage{algorithm}
\usepackage{algorithmic}
\usepackage{hyperref}

\usepackage{caption}
\usepackage{subcaption}
\usepackage{helvet}
\usepackage{courier}
\usepackage{color}
\usepackage{amsmath}
\usepackage{amsfonts}
\usepackage{commath}

\numberwithin{equation}{subsection} %{section}

\icmltitlerunning{Grouped Recurrent Convolutional Layers for Multivariate Time Series}

\begin{document} 

\twocolumn[
\icmltitle{Grouped Convolutional Neural Networks for Multivariate Time Series}

% It is OKAY to include author information, even for blind
% submissions: the style file will automatically remove it for you
% unless you've provided the [accepted] option to the icml2017
% package.
\icmlsetsymbol{equal}{*}
\begin{icmlauthorlist}
\icmlauthor{Subin Yi}{unist}
\icmlauthor{Janghoon Ju}{unist}
\icmlauthor{Man-Ki Yoon}{uiuc}
\icmlauthor{Jaesik Choi}{unist}
\end{icmlauthorlist}
\icmlaffiliation{unist}{Ulsan National Institute of Science and Technology, Ulsan, 44919, Republic of Korea}
\icmlaffiliation{uiuc}{University of Illinois at Urbana-Champaign,
			Urbana, IL 61801}
\icmlcorrespondingauthor{Jaesik Choi}{jaesik@unist.ac.kr}

% You may provide any keywords that you 
% find helpful for describing your paper; these are used to populate 
% the "keywords" metadata in the PDF but will not be shown in the document
\icmlkeywords{deep learning, time series analysis, group convolutional neural networks}
\vskip 0.3in
]

\printAffiliationsAndNotice{}

%%%%%%%%%%%%%%%%%%%%%%%%%%%%%%%%%%%%%%%%%%%%%
% up to eight pages long, not including references.                              %
% up to ten pages when references and acknowledgment are included. %
%%%%%%%%%%%%%%%%%%%%%%%%%%%%%%%%%%%%%%%%%%%%%

\begin{abstract} 
Analyzing multivariate time series data is important for many applications such as automated control, fault diagnosis and anomaly detection. One of the key challenges is to learn latent features automatically from dynamically changing multivariate input. In visual recognition tasks, convolutional neural networks (CNNs) have been successful to learn generalized feature extractors with shared parameters over the spatial domain.
%and therefore has been a popular tool for the image processing tasks. 
However, when high-dimensional multivariate time series % with an unknown correlated structure 
is given, designing an appropriate CNN model structure becomes challenging because the kernels may need to be extended through the full dimension of the input volume.
%Thus, it is hard to learn CNN features and justify the trained  features. 
To address this issue, we present two structure learning algorithms for deep CNN models. Our algorithms exploit the covariance structure over multiple time series to partition input volume into groups. The first algorithm learns the group CNN structures explicitly by clustering individual input sequences. The second algorithm learns the group CNN structures implicitly from the error backpropagation. % kernels for We present two algorithms to Our G-RCLs exploit the correlation over multiple time-series and construct shared convolutional features by clustering time series.% Moreover, we propose an iterative learning algorithm to find optimal structure for G-RCNN.
%Moreover, we introduce two implementations of G-RCLs: hard and soft.
In experiments with two real-world datasets, we demonstrate that our group CNNs outperform existing CNN based regression methods.   
% In experiments with data collected from a quadcopter, a drone with four wings.
% Also, we demonstrate that a Recurrent Convolutional Neural Network (RCNN) which includes grouped recurrent convolutional layer (G-RCL) outperforms a general RCNN to detect abnormal events in complex systems.
\end{abstract}

\section{Introduction}\label{sec:intro}

Advances in computing technology has made many complicated systems such as automobile, avionics, and industrial control systems more sophisticated and sensitive. 
Analyzing multiple variables that compose such systems accurately is therefore becoming more important for many applications such as automated control, fault diagnosis, and anomaly detection.

In complex systems, one of the key requirements is to maintain integrity of the sensor data so that it can be monitored and analyzed in a trusted manner. Previously, sensor integrity has been analyzed by feedback controls \cite{SecureEstimationWIthIntegrityAttacks, AttackResilientStateEstimator} and nonparametric Bayesian methods \cite{SensorPlacementGPKrause2008}. However, regression models based on control theory and nonparametric Bayesian are highly sensitive to the  model parameters. Thus, finding the best model parameter for the regression models is challenging with high-dimensional multivariate sequences.

Artificial neural network models also have been used to handle multivariate time series data. Autoencoders \cite{AutoBourlard1988,AutoencodersZemel1994} train model parameters in an unsupervised manner by specifying the same input and output values. Recurrent neural networks (RNN) \cite{RNN:original} and long-short term memory (LSTM) \cite{RNN:LSTM} represent changes of time series data by learning recurrent transition function between time steps. Unfortunately, existing neural network models for time series data assume fully connected networks among time series under the Markov assumption. Thus, such models are often not precise enough to address high-dimensional multivariate regression problems.

To address this issue, we present two structure learning algorithms for deep convolutional neural networks (CNNs). Both of our algorithms partition input volume into groups by exploiting the covariance structure for multiple time series so that the input CNN kernels process only one of the grouped time series. Due to this partitioning of the input time series, we can avoid the CNN kernels being extended through the full dimension.
 In this reason, we denote the CNN models as \emph{Group CNN (G-CNN)} which can exploit latent features from multiple time-series more efficiently by utilizing structural covariance of the input variables. 

The first structure learning algorithm learns the CNN structure explicitly by clustering input sequences with spectral clustering \cite{TutorialonSpectralClustering}. The second algorithm learns the CNN structures implicitly with the error backpropagation which will be explained in Section~\ref{ssec:NNwithSoftClustering}.
%Especially, recurrent neural network (RNN) based models are the popular choices due to its effectiveness in extracting temporal features.

%We use a deep learning model to learn the unknown dynamics of multiple time series and estimate one or more temporal variables. By comparing the collected sensor signals and values estimated by the model, the artificial neural networks decide if the sensor is trustworthy. This approach does not require the system be formulated and can be generalized in a variety of fields that consist of multiple temporal variables. 
%In this paper, we introduce a new deep neural network architecture, Grouped RCNN (G-RCNN), to analyze multiple time series. \manki{In this paper, we introduce Grouped RCNN (G-RCNN), a novel deep neural network architecture, to analyze multiple time series IN WHAT SENSE?.}

Our model design principle is to reduce model parameters by sharing parameters when necessary. In multivariate time series regression tasks, our hypotheses on the parameter sharing scheme (or parameter tying) are as follow:  (1) convolutions on a group of correlated signals are more robust to signal noises; and (2) convolutions operators on groups of signals are more feasible to learn when a large number of time series is given. In experiments, we show that G-CNN make the better predictive performance on challenging regression tasks compared to the existing CNN based regression models.

\section{Background}

\subsection{Convolutional Neural Network}
\begin{figure}[t]
  \centering
  \includegraphics[width=\linewidth]{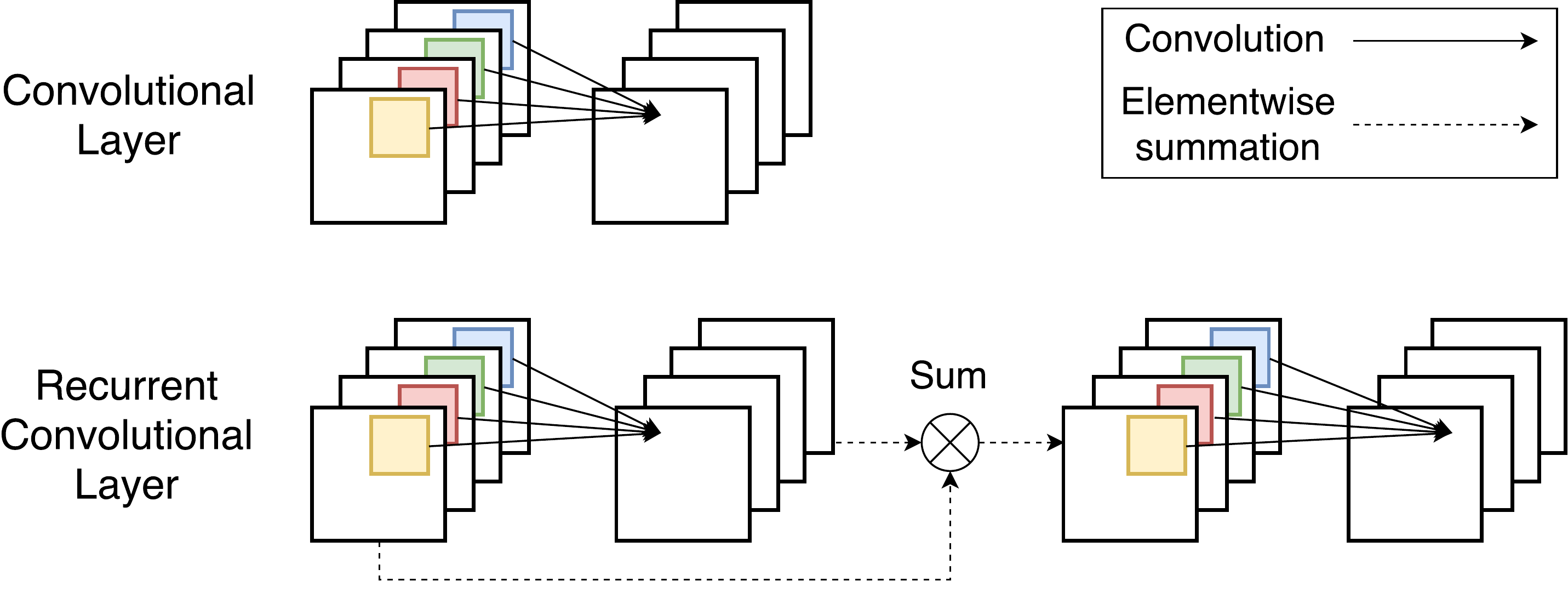}
  \caption{Building blocks of CNN and RCNN.}
  \label{layers1}
\end{figure} 
A convolutional neural network (CNN) is a multi-layer artificial neural network that has been successful recognizing visual patterns. The most common architecture of the CNNs is a stack of three types of multiple layers: convolutional layer, sub-sampling layer, and fully-connected layer. Conventionally, A CNN consists of alternate layers of convolutional layers and sub-sampling layers on the bottom and several fully-connected layers following them. 

First, an unit of a convolutional layer receives inputs from a set of neighboring nodes of the previous layer similarly with animal visual cortex cell. The local weights of convolutional layers are  shared with the nodes in the same layer. Such local computations in the layer reduce the memory burden and improve the classification performance.

None-linear down-sapling layer, which is the second type of CNN layers, is another important characteristic of CNNs. The idea of local sub-sampling is that once a feature has been detected, its location itself is not as important as its relative location with other features. By reducing the dimensionality, it reduces the local sensitivity of the network and computational complexity \cite{LeCun95,LeCun98}. 

The last type of the layers is the fully-connected layer. It computes a full matrix calculation with all activations and nodes same as regular neural networks. After convolutional layers and sub-sampling layers extract features, fully-connected layers implements reasoning and gives the actual output. Then the model is trained in the way minimizing the error between the actual output of the model and the target output values by backpropagation method.

CNN has been very effective for solving many computer vision problems such as classification \cite{LeCun98,Krizhevsky12}, object detection and semantic segmentation \cite{Ren15,Long15}. It has been also applied to other problems such as natural language processing \cite{Kalchbrenner14,Collobert08}. Recently, variants of CNN are applied to analyzing various kinds of time-series such as sensor values and EEG (electroencephalogram) signals \cite{Yang15,Ordóñez16}.

\subsection{Recurrent Convolutional Neural Network (RCNN)}\label{ssec:RCNN}
 %CNN and its variations have become very popular model due to their effectiveness of extracting features from locally correlated data. 
 Recurrent Convolutional Neural Network (RCNN) is a type of CNN with its convolutional layers being replaced with recurrent convolutional layers. It improves the expressive power of the convolutional layer by exploiting multiple convolutional layers that share the parameters. RCNN has been applied to not only the image processing problem \cite{Liang15,Pinheiro14} but also other tasks that require temporal analysis \cite{Lai15}. 
RCNN can effectively extract invariant features in the temporal domain regarding the time-series data as a 2-dimensional data with one of the dimensions is one. with one of the dimensions is 1, RCNN can effectively extract invariant features in the temporal domain.

\subsubsection{Recurrent Convolutional Layer}
Recurrent Convolutional Layer (RCL), which is the most representative building block of an RCNN, is the composition of $l$ intermediate convolutional layers that shares the same parameters.
The first convolutional layer of an RCL carries the  convolution on the input $\textbf{x}$, resulting in the output $\sigma(W*\textbf x)$  where $W$ is the convolutional filter, * is a convolution operator, and $\sigma(\cdot)$ is an activation function. Then the next convolutional layer recursively processes the summation of the original input and the output of the previous layer, $\textbf x + \sigma(W*\textbf x)$, as an input. After some iterations of this process, an RCL gives the result of the final intermediate convolutional layer as its output. 

During the error backpropagation, the parameters are updated $l$ times. In each update, the parameters are changed to fix the error made by itself from the previous layer. %It can also learn the dependencies between the inputs as it takes the result of the convolution.

RCL can also be regarded as a skip-layer connection \cite{He15,Intrator01}. Skip-layer connection represents connecting layers skipping intermediate layers as in Figure~\ref{layers1}'s RCL. The main motivation is that the deeper networks show a better performance in many cases but they are also harder to train in actual applications due to vanishing gradients and degradation problem \cite{Bengio94,Glorot10}.

\cite{He15} designed such layers with skip-layer connection, named as residual learning. The idea was that if one can hypothesizes that multiple nonlinear layers can estimate an underlying mapping $H(x)$, it is equivalent to estimating an residual function $F(x) := H(x)-x$. If the residual $F(x)$ is approximately a zero mapping, $H(x)$ is an optimal identity mapping.

\subsection{Spectral Clustering}
\label{spectralclustering}
 The goal of clustering data points $\mathbf{x}_1,...,\mathbf{x}_N$ is to partition the data points into some groups such that the points in the same group are similar and points in different groups are dissimilar in a certain similarity measure $s_{ij}$ between $x_i$ and $x_j$. Spectral clustering is the clustering method that solves this problem from the graph-cut point of view. 
 
From the graph-cut point of view, data points are represented as a similarity graph $G=(V, E)$. 
Let $G$ be a weighted undirected graph with the vertex set $V=\{v_1,...,v_N\}$ where each vertex $v_i$ represents a data point $x_i$ and the weighted adjacency matrix $W=\{w_{ij}| i,j=1,...,N\}$ where $w_{ij}$ represents the similarity $s_{ij}$. 
Let the degree of a vertex $v_i \in V$ be $d_i = \sum^n_{j=1}w_{ij}$ and define a degree matrix $D$ as the diagonal matrix with the degrees $d_1,...,d_N$ on the diagonal.

Then, clustering can be reformulated to find a partition of the graph such that the edges between different groups have very low weights and the edges within a group have high weights.

One of the most intuitive way to solve this problem is to solve the min-cut problem \cite{NormalizedCuts}. Min-cut problem is to choose a partition $A_1,...,A_K$ for a given number $K$ that minimizes the equation (\ref{eqn:cut}) given as:
\begin{align}
	\mbox{cut}(A_1,...,A_K) &:= \frac{1}{2}\sum_{i=1}^K \mbox{link}(A_i, \bar{A}_i) \label{eqn:cut}\\
    \textnormal{s.t.  } \mbox{link}(A,B)&:=\sum_{i\in A, j\in B}w_{ij}\textnormal{ for disjoint } A,B\subset A.\label{eqn:link}
\end{align}
Here, 1/2 is introduce for normalizing as otherwise each edge will be counted twice. The algorithm of \cite{NormalizedCuts} explicitly requests the sets be large enough where the size of a subset $A\subset V$ is measured by:
\begin{align}
	\mbox{vol}(A):=\sum_{i\in A}d_i
\end{align}

Then, find the following normalized cut, Ncut:
\begin{align}
	\mbox{Ncut}(A_1,...,A_K) :=\frac{1}{2}\sum_{i=1}^K \frac{\mbox{link}(A_i,\bar A_i)}{\mbox{vol}(A_i)}
\end{align}
The denominator of the Ncut tries to balance the size of the clusters and the numerator finds the minimum cut of the given graph. Then to find the partition $A_1,...,A_K$ is same as to solve the following optimization problem:
\begin{align}
	&\min_{H\in \mathbb{R}^{NxK}} trace(H^TLH)\\
    s.t. \hspace{3em} &H_{ij} :=\begin{cases}
                        \frac{1}{\sqrt{vol(V_j)}}, v_i\in V_j\\
                        0, otherwise
                    \end{cases}\\
        &L := D -W
\end{align}
As $h^T_iLh_i {=} \mbox{cut}(A_i,\bar A_i)/\mbox{vol}(A_i)$, $H^TH{=}I$, and $h_i^TDh_i{=}1$ where the indicator vector $h_j$ is the j-th column of the matrix H. 

Unfortunately, introducing the additional term to the min-cut problem has proven to make the problem NP-hard \cite{MinCutandGraphBisection} so \cite{TutorialonSpectralClustering,LearningSpectralClustering} solves the relaxed problem, which gives the solution H that consists of the eigenvectors corresponding to the K smallest eigenvalues of the matrix $L_{rw}:=D^{-1}L$ or the K smallest generalized eigenvectors of $Lu=\lambda Du$. 

%Given the similarity matrix $S\in\mathbf R^{nxn}$ and the number $k$ of clusters to construct, the spectral clustering algorithm of \cite{NormalizedCuts} constructs a similarity graph G with a adjacency matrix W. Then it computes the first k generalized eigenvectors $u_1,...,u_K$ of the generalized eigenproblem $Lu=\lambda Du$ and build a matrix $U \in \mathbf R^{NxK}$ letting $u_1,...,u_K$ be its columns. Finally, n rows $\{y_i\in \mathbf R^K|i=1,...,N\}$ of the matrix U are grouped into clusters $C_1,...,C_K$ with K-means clustering algorithm and returns the clusters $A_1,...,A_K$ where $A_i=\{j|y_j\in C_i\}$.

 Given an $N$x$N$ similarity matrix, the spectral clustering algorithm runs eigenvalue decomposition (EVD) on the graph Laplacian matrix and the eigenvectors corresponding to the K smallest eigenvalues are clustered by a clustering algorithm representing the graph vertices. The K eigenvectors are also the eigenvectors of the similarity matrix whereas corresponding K largest eigenvalues, which can be considered as an encoding of the graph similarity matrix.
%In the work of \cite{GraphClustering}, they argued that the spectral clustering and AutoEncoder are similar when they are used to solve graph clustering problems (i.e. the autoencoder tries to reconstruct the normalized similarity matrix with the error calculated by Frobenuis norm).

\section{Grouped Time Series}\label{sec:GroupedTimeSeries}
In this section, we present two algorithms build group CNN structure. The first method builds the group structure explicitly from Spectral clustering. The second method build the group structure through the error backpropagation.

\subsection{Learning the Structure by Spectral Clustering}\label{ssec:model-explicit}
\begin{figure}[t!]
  \centering
  \includegraphics[width=\linewidth]{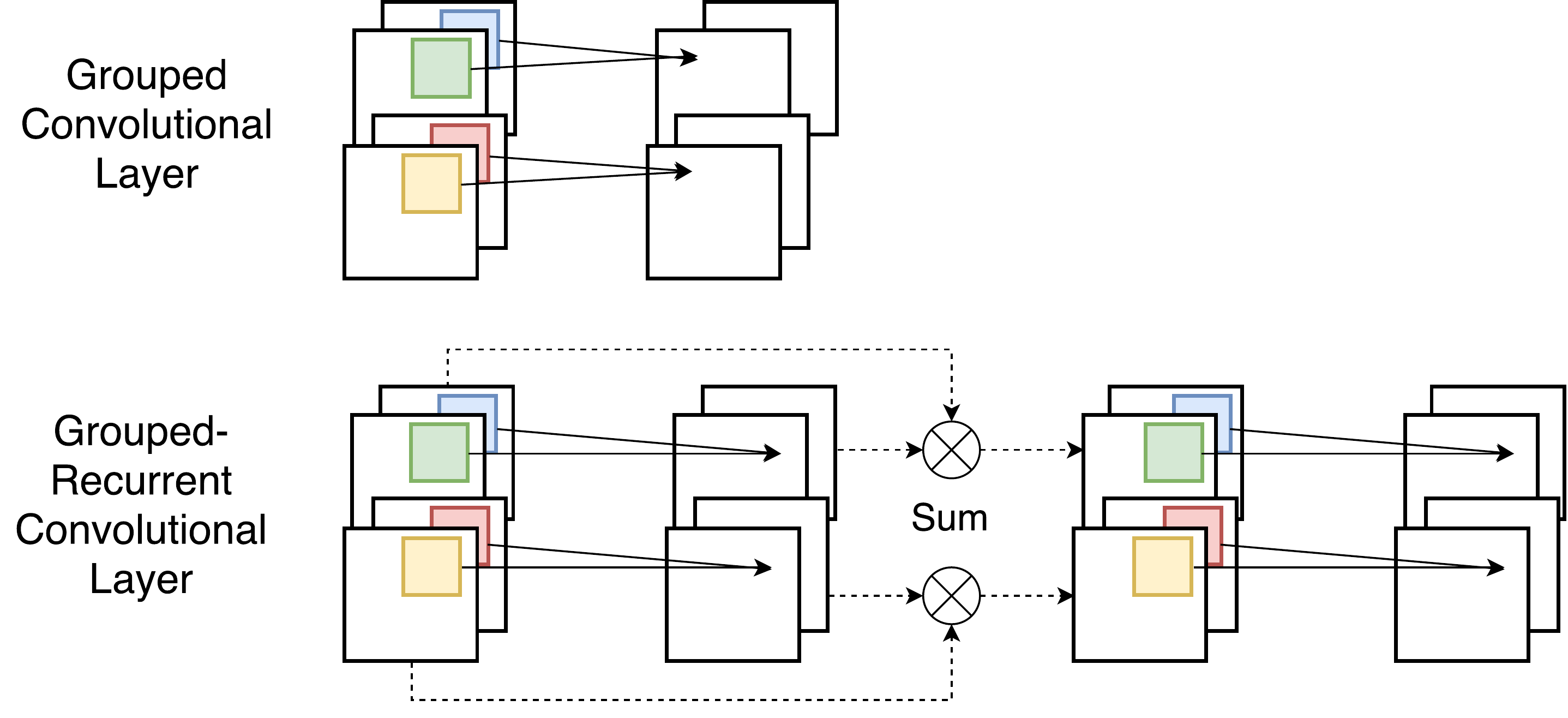}
  \caption{Grouped layers for CNN and RCNN.}
  \label{fig:layers2}
\end{figure} 

\begin{figure*}[t!]
    \begin{subfigure}{\textwidth}
        \includegraphics[width=\textwidth]{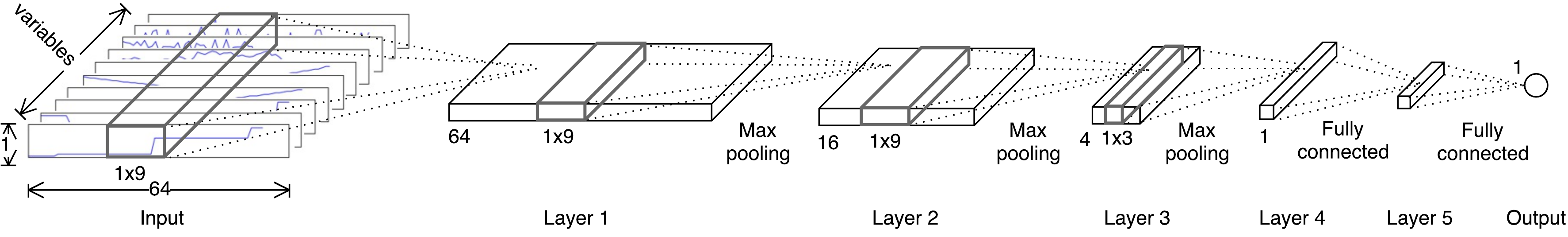}
		\caption{Convolutional Neural Network (CNN).}
		%\label{RCNN}
    \end{subfigure}
    \begin{subfigure}{\textwidth}
		\includegraphics[width=\textwidth]{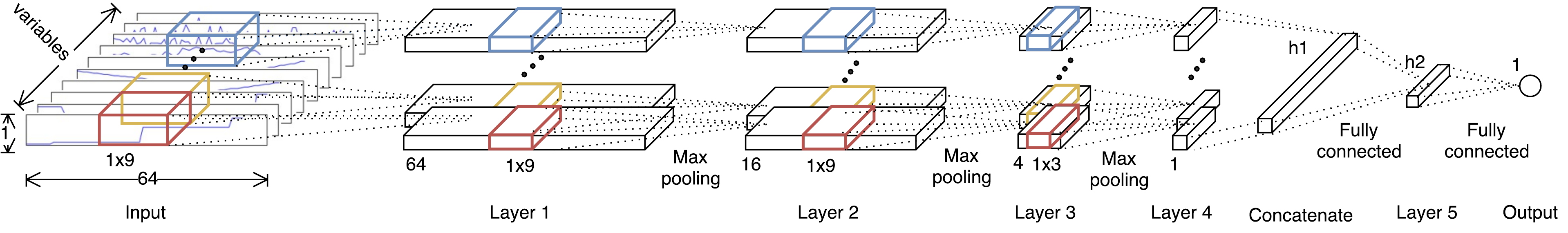}
		\caption{CNN with grouped convolutional layers.}
		%\label{R-RCL}
    \end{subfigure}
    \caption{Comparison of the general CNN model and CNN with grouped layers.}
    \label{fig:RCNN_compare}
    %Illustrations of an  general RCNN model and an RCNN with grouped RCLs.
\end{figure*}
Our group CNN structure receives both the input variables $\mathbf X = [\mathbf x_1, ..., \mathbf x_N]$ and their cluster information $C = [c_1,...,c_N]$ where $c_i$ represents the membership of the variable $x_i$. Unlike usual convolutional layers, the grouped convolutional layers divide the input volume based on the cluster membership $c_i$ and performs the convolution operations over the input variables that belong to the same cluster as described in the Figure~\ref{fig:layers2}. Formally, the k-th group of the layer H is defined as:
\begin{align}
\label{eqn:gCNN}
	% H^k = \sigma\big(\sum_{i=\{1,...,N | c_i=k\}} W^k\cdot\mathbf{x}_i + b^k\big)
    H^k = \sigma\big(\sum_{i} W^k\cdot\mathbf{x}_i + b^k\big)
\end{align}
where $(\cdot)$ is the convolution operation, $i\in \{j| j\in\{1,...,N\}, c_j=k\}$, $W^k$ is the weight matrix, and $b_k$ is the bias vector of the k-th group.

As in the CNN models, the input variables $X$ are processed throughout multiple grouped convolutional layers and sub-sampling layers, flattened into one-dimensional layer followed by fully-connected layers, and produces the output $\mathbf y=\{y_1,...,y_P\}$ (Figure~\ref{fig:RCNN_compare}). 

 Given the target output $\mathbf t=\{t_1,...,t_P\}$, we can also train this model using gradient descent solving the optimization problem: %by minimizing the error between $\mathbf y$ and the target output $\mathbf t$.
\begin{align}\label{eqn:errfunc}
	\min_\theta \sum_{i=1}^{p}(y_i-t_i)^2
\end{align}
with respect to the trainable parameter $\theta=\{W, b\}$. The error is backpropagated to each group separately, training the CNN structure explicitly.

This model requires significantly less number of parameters compared to the vanilla CNN model. For example, to process 100 input variables producing 100 output channel, existing CNN model needs 100 kernels of size (width, height, 100). However, if the input variables consist of 5 clusters, each with 20 variables, it requires 5x20 kernels of size (width, height, 20), which is 5 times less than the vanilla model. It could make the CNN model more compact by eliminating redundant parameters. 

\subsection{Neural Networks with Clustering Coefficient} \label{ssec:NNwithSoftClustering}

\begin{figure}[t!]
  \centering
  \includegraphics[width=0.8\linewidth]{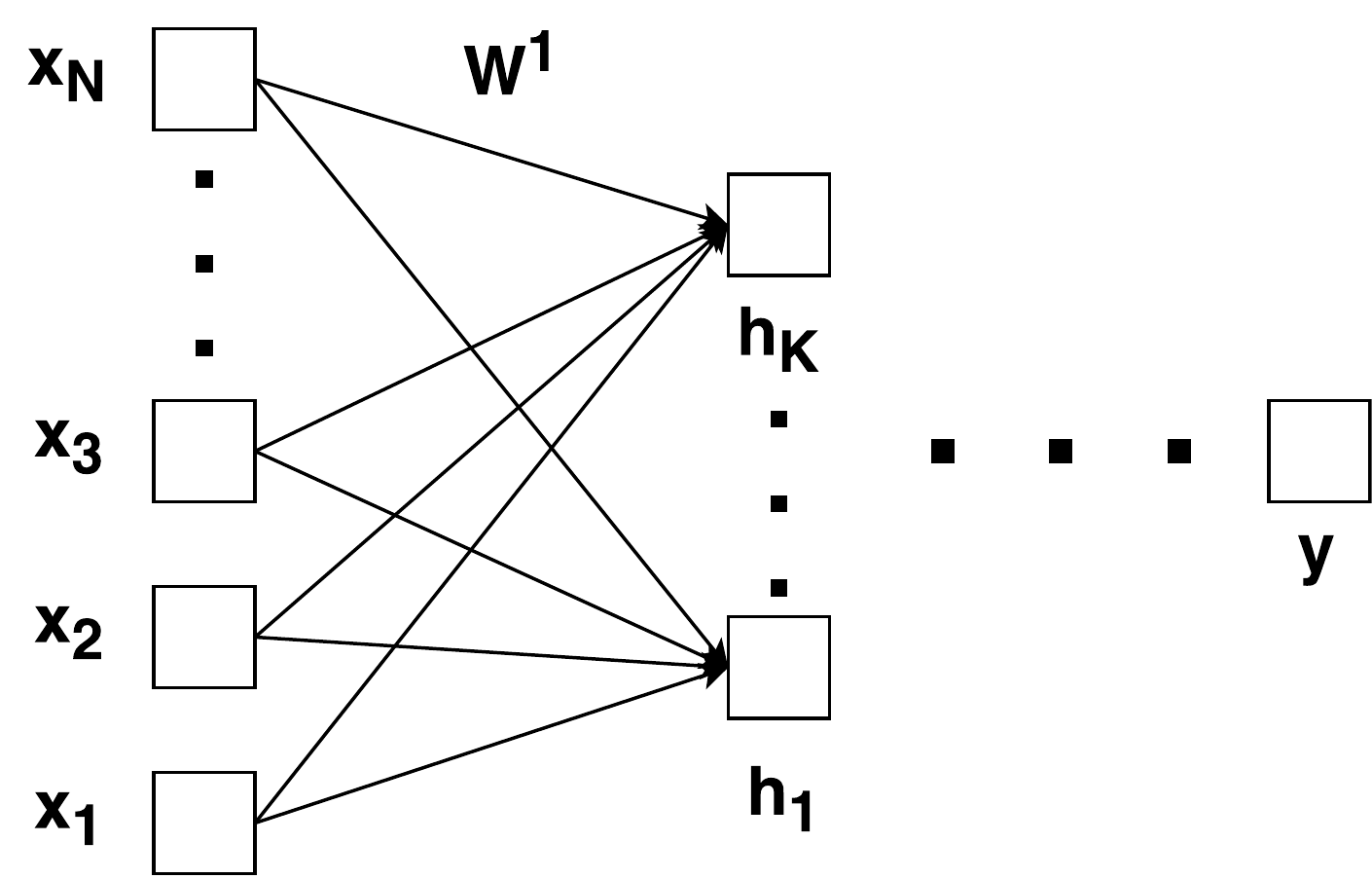}
  \caption{Example of a neural network that receives grouped variables.}
  \label{fig:groupedVarialbes}
\end{figure} 

 Assuming that the input time series are correlated with each other, we group those variables explicitly to make use of such correlations as CNN utilizes local connectivity of an image. It can be considered to find the local connectivity and correlations within channels of CNN.
 
 Given an input data $X$ which consists of N variables where each variables are $D$ dimensional real valued vectors, i.e. $X=[\mathbf{x}_1, ... ,\mathbf{x}_N]$, we wish to group these variables into K clusters introducing a matrix $U=[u_{i,j}; i\in\{1,...,N\}, j\in\{1,...,K\}]$ where $u_{i,j}\in[0,1], \sum_j^K u_{i,j} = 1$ whose element $u_{i,j}$ is the clustering coefficient which represents the portion of the $j$-th cluster takes for the variable $\mathbf x_i$ as in multinomial distribution. In this paper, we use boldface letters to represent D-dimensional real valued column vectors.%$\mathbf{x}_i\in \R ^D, i\in[N]$.
 Then, $\mathbf{h}_j$, the node that represents the j-th cluster is defined as :
\begin{align}
	\mathbf{h}_j = \sigma(\sum_i^N u_{i,j} \mathbf x_i^T \cdot 
    					  \mathbf W^1_{i,j} +\mathbf b^1_j )
\end{align}
where $\mathbf W^1_{i,j}$ is the i-th row's j-th column of the weight matrix $W^1$ of size $N$x$K$, $\mathbf b^1_j$ is the bias, and $\sigma(\cdot)$ is the activation function. By multiplying $u_{i,j}$, variables can proportionally participate to each cluster.

 Suppose that an example of two layered neural network is given as shown in the Figure~\ref{fig:groupedVarialbes}. The output node $\mathbf y$ is also defined as :
 \begin{align}
 	\mathbf y = \sigma(\sum_j^K \mathbf h_j^T \cdot \mathbf W^2_{j,1} +\mathbf b^2_1 ).
 \end{align}
Given the true target value $\mathbf t$, this network can be trained by gradient descent method solving the below optimization problem:
\begin{align}
	\min_{W, U, B}& Err,\\
    Err :=& \frac{1}{2}(\mathbf y-\mathbf t)^2
\end{align}
Assuming linear activation function, gradient of the $Err$ with respect to $u_{i,j}$ is :
\begin{align}
	\frac{\partial Err}{\partial u_{i,j}} &= 
    	\frac{\partial Err}{\partial \mathbf y} \frac{\partial \mathbf y}{\partial u_{i,j}} \nonumber\\
        &= \frac{\partial Err}{\partial \mathbf y} \frac{\partial \mathbf y}{\partial \mathbf h_j}
        	\frac{\partial \mathbf h_j}{\partial \mathbf u_{i,j}} \nonumber\\
            &= \frac{\partial Err}{\partial \mathbf y} (\frac{\partial \mathbf h^T_j}{\partial u_{i,j}} 
            							+ \sum_{j'}^K\mathbf{I}\{j'\neq{j}\}\frac{\partial\mathbf h^T_{j'}}{\partial u_{i,j'}}
                                        \frac{\partial u_{i,j'}}{\partial u_{i,j}})
                                        \frac{\partial \mathbf h_j}{\partial \mathbf u_{i,j}} \nonumber\\
        &= (\mathbf y - \mathbf t) 
        	(\mathbf K\mathbf x^T_i\cdot\mathbf W^1_{i,j} - \sum_{j}^K\mathbf x^T_i\cdot \mathbf W^1_{i,j})
            (\mathbf x^T_i\cdot\mathbf W^1_{i,j})
\end{align}
where $j'$ is the cluster out of the j-th cluster and $\mathbf{I}$ is an indicator function. Intuitively, the parameter update $u_{i,j}$ includes (K times of loss from the j-th cluster - loss from all clusters ; $\mathbf Kx^T_i\cdot\mathbf W^1_{i,j} - \sum_{j}^K\mathbf x^T_i\cdot \mathbf W^1_{i,j})$, $u_{i,j}$ value that gives smaller loss increase finding the optimal values while minimizing the error by the gradient descent method.

\begin{figure}[h!]
  %\centering
  \includegraphics[width=0.7\linewidth]{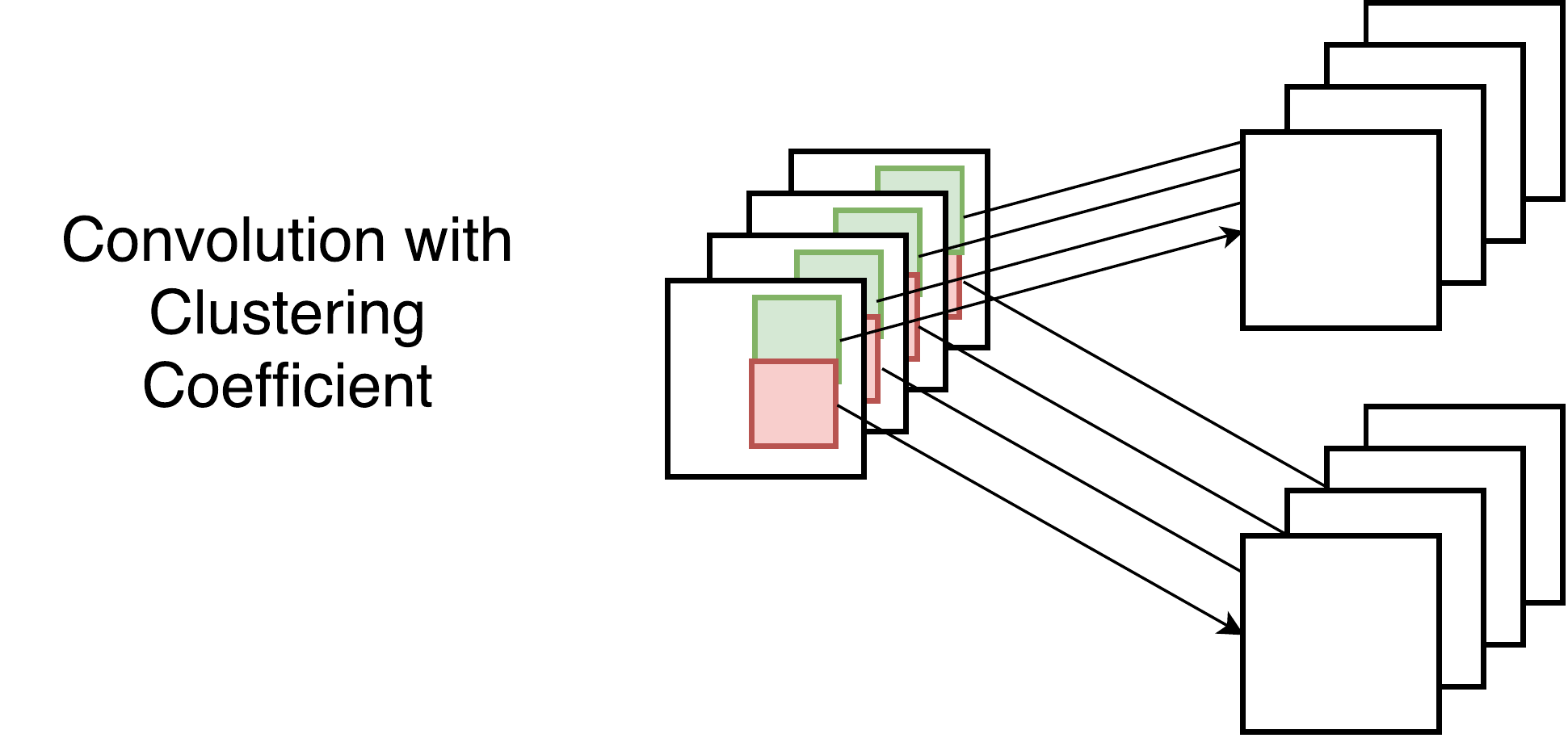}
  \caption{Convolutional layer with clustering coefficient.}
  \label{fig:layer3}
\end{figure} 
To implement the clustering coefficient to out model, we added a new layer which works as the Figure~\ref{fig:layer3} on the bottom of the model, before the layer 1 of the Figure ~\ref{fig:RCNN_compare} (b). This layer receives $N$ input variables and computes channel-wise convolution throughout the variables using the same weight and  bias in the group (group parameter sharing). This channel-wise convolution is repeated for K groups with different parameters for each groups. Therefore, the i-th channel of the k-th group is defined as:
\begin{align}\label{eqn:gCNN-soft}
	h^K_i = \sigma(u_{i,k} W^k\cdot \mathbf x_i + b^k)
\end{align}
Then the output is processed by the same process with the model from the Section~\ref{ssec:model-explicit} with explicit clustering.

% \subsection{Grouping by Implicit Learning}\label{ssec:model-implicit}
% Implementation of the Section~\ref{ssec:NNwithSoftClustering}.

\section{Related Work}
 Recently, deep learning methods are making good results in solving a variety of problems such as visual pattern recognition, signal processing, and others. Consequently there has been researches for applying those methods to analyzing complex multivariate systems.

Neural networks that are composed of fully-connected layers only are not appropriate for handling sequential data since they need to process the whole sequence of input. More specifically, such networks are too inefficient in terms of both memory usage and learning efficiency.

One of the popular choices for processing time-series is a recurrent neural network (RNN). An RNN \cite{RNN:tutorial,RNN:dynamical-system} processes sequential data with recurrent connections to represent transition models over time. so that it can store temporal information within the network. RNN models have been successfully used for processing sequential data \cite{MultiscaleTemporalStructure, TrainingRNN, ClockworkRNN}. \cite{RNNhydrologicalforecasting} used dynamic RNN to forecast nonstationary hydrological time-series and 
\cite{LSTManomalydetection} used stacked LSTM network as a predictor over a number of time steps and detected anomalies that has high prediction error in time series. 

Convolutional neural network (CNN) is also commonly used to analyze temporal data.\cite{CNNspeechrecognition} used CNN for speech recognition problem and \cite{CNNtimeseriesclassification} proposed multi-channels deep convolutional neural network for multivariate time series classification.

Recurrent Convolutional Neural Network (RCNN), which can be considered as a variant of a CNN, is recently proposed and shows state-of-the-art performance on classifying multiple time series \cite{Pinheiro14,Liang15,Lai15}. When a small number of time series is given, multiple signals can be handled individually in a straightforward manner by using polling operators or fully connected linear operators on signals. However, it is not clear how to model the covariance structure of large number of multiple sequences explicitly for deep neural network models.

\section{Experimental Results}
In experiments, we compare the regression performance of several CNN-based models on two real-world high-dimensional multivariate datasets, groundwater level data and drone flight data. Groundwater data and drone data respectively have 88 and 148 variables. 
\begin{table*}[th!]
  %\centering
  \begin{tabular}{ |c|c||c|c|c|c|c|c|c|  }
     \hline
     %\multicolumn{3}{|c|}{Country List} \\
     %\hline
     & Model  & Input & Layer 1 & Layer 2 & Layer 3 & Layer 4 & Layer 5 & Output\\
     \hline
     \hline
     					& CNN					& 1x64x87 & 1x64x500 & 1x16x500& 1x4x500& 1x1x500& 100 & 1\\
     					& RCNN\textsuperscript{3} & 1x64x87 & 1x64x500& 1x16x500& 1x4x500& 1x1x500& 100 & 1\\
                        \cline{2-9}
     Water\textsuperscript{1} & CNN \& exp 	& 1x64x87 & (1x64x100)5\textsuperscript{4}& (16x100)5& (4x100)5& (1x100)5& 100&1\\
     					& RCNN\textsuperscript{3} \& exp & 1x64x87 & (1x64x100)5& (16x100)5& (4x100)5& (1x100)5& 100&1\\
     					& CNN \& coeff 	\textsuperscript{6}	& 1x64x87 & (1x64x100)5& (16x100)5& (4x100)5& (1x100)5& 100&1\\
     					& RCNN \& coeff & 1x64x87 & (1x64x100)5& (16x100)5& (4x100)5& (1x100)5& 100&1\\
     \hline
     \hline
     		   & CNN				   & 1x64x147  & 1x64x750& 1x16x750& 1x4x750& 1x1x750& 200& 1\\
     		   & RCNN\textsuperscript{3} & 1x64x147 & 1x64x750& 1x16x750& 1x4x750& 1x1x750& 200& 1\\
     \cline{2-9}
     Drone\textsuperscript{2} & CNN \& exp   & 1x64x147 & (1x64x50)15\textsuperscript{5}& (1x16x50)15& (1x4x50)15& (1x1x50)15& 200& 1 \\
     		   & RCNN\textsuperscript{3} \& exp & 1x64x147 & (1x64x50)15& (1x16x50)15& (1x4x50)15& (1x1x50)15& 200& 1 \\
     		   & CNN \& coeff \textsuperscript{6}	& 1x64x147 & (1x64x50)15& (1x16x50)15& (1x4x50)15& (1x1x50)15& 200& 1\\
               & RCNN \& coeff & 1x64x147 & (1x64x50)15& (1x16x50)15& (1x4x50)15& (1x1x50)15& 200& 1\\
     \hline
  \end{tabular}
  \raggedbottom{\textsuperscript{1}\footnotesize{Groundwater Dataset.}}
  \raggedbottom{\textsuperscript{2}\footnotesize{Drone Dataset.}}
  \raggedbottom{\textsuperscript{3}\footnotesize{Layer 1, Layer 2, Layer 3 are RCLs with iteration 2.}}
  \raggedbottom{\textsuperscript{4}\footnotesize{K=5.}}
  \raggedbottom{\textsuperscript{5}\footnotesize{K=15.}}\\
  \raggedbottom{\textsuperscript{6}\footnotesize{Models with clustering coefficient has an additional layer (Figure~\ref{fig:layer3}) before the Layer 1.}}
  \caption{Architecture of tested deep CNN models.}
  \label{table:architecture}
\end{table*}

\subsection{Settings}
%\FloatBarrier
To evaluate the regression performance, we picked one of the variables, say $x_p$, from the dataset randomly and constructed the variable's values as a target at time t, $y = x_p(t)$, by seeing its correlated variables' values from time $t-T$ to $t$ without including variable, i.e. $X = \cup_{i, i\neq p} [x_i(t-T),...,x_i(t)]$.
 
 We trained our models with $90\%$ of the whole dataset and tested on the other $10\%$. Then, the regression performance were compared with other regression models: linear regression, ridge regression, CNN and RCNN. %The architecture of the tested models are shown in the Table~\ref{table:architecture}.
%Performance was measured by evaluating how accurately it can construct a variable with its correlated variables without itself. To measure the regression performance of tested models, normalized root mean square error (NRMSE) was used:
%We used the fore part $90\%$ of the total dataset to train our model and tested on the other $10\%$. 
The regression performance was measured on the scale of the standardized root mean square error (SRMSE), which is defined as the equation~\ref{eq:SRMSE} when $\bar t$ is the mean value of the target vector $\mathbf t$.
\begin{align}
 	\textnormal{SRMSE} &= \frac{\sqrt{\Sigma_{t=1}^{N}(t_t-y_t)^2/N}}{SE}\label{eq:SRMSE}\\
    SE &= \sqrt{\frac{1}{N}\sum_i(t_i-\bar t)^2}
\end{align}
 %And true positive rate (TPR), true negative rate (TNR), false positive rate (FPR), and false negative rate (FNR) were measured to evaluate the anomaly detection performance where the true positive 

%\subsection{Groundwater Data}\label{ssec:groundwater}
\subsection{Datasets}\label{ssec:Dataset}
\subsubsection{Groundwater Data}
\label{sssec:groundwater}

\begin{figure}[h]
	\centering
    \begin{subfigure}{0.9\linewidth}
    	\centering
        \includegraphics[width=0.9\linewidth]{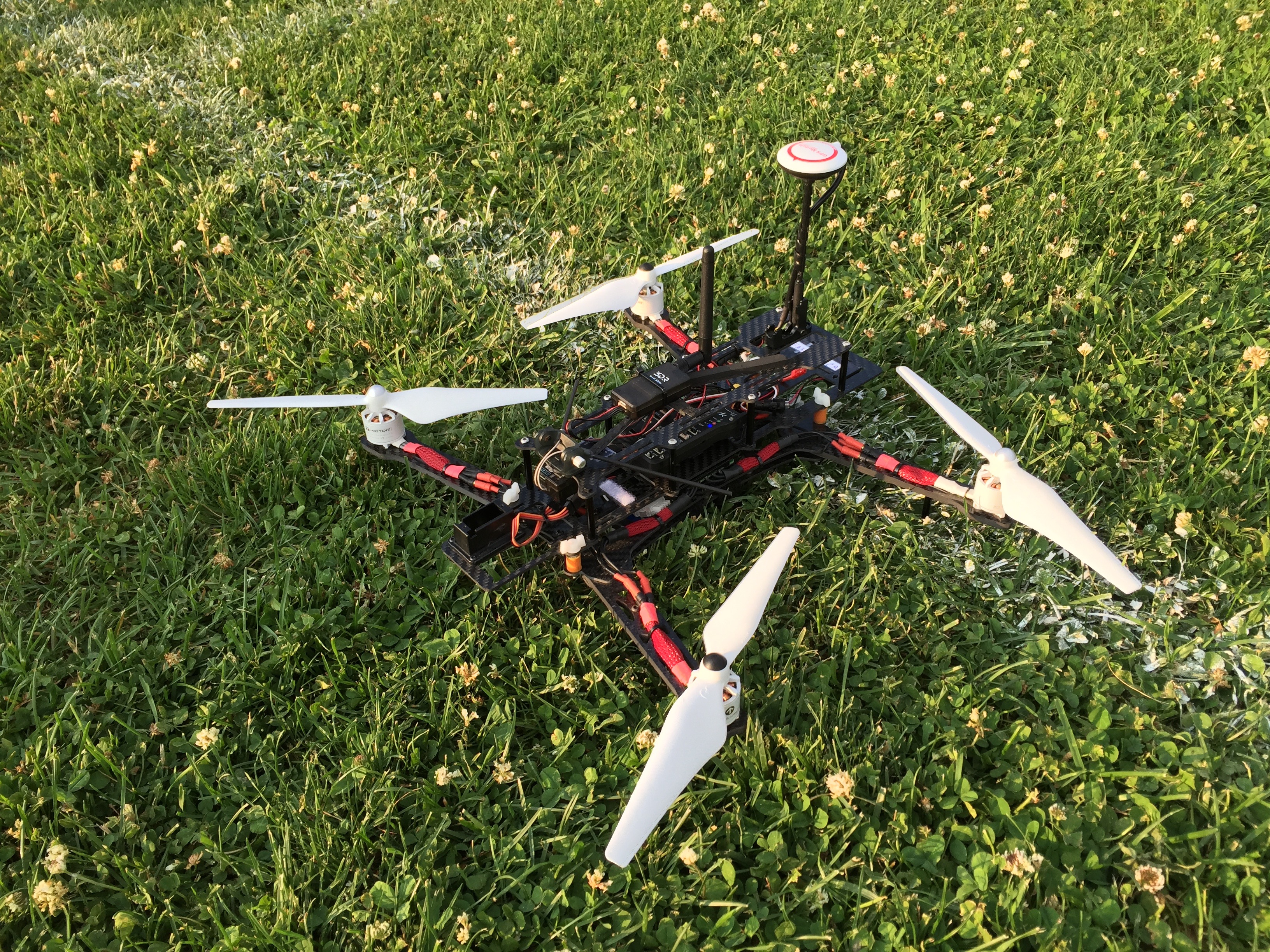}
		\caption{quadcopter used for the test.}
		\label{fig:path2}
    \end{subfigure}
    \begin{subfigure}{0.9\linewidth}
    	\centering
        \includegraphics[width=0.9\linewidth]{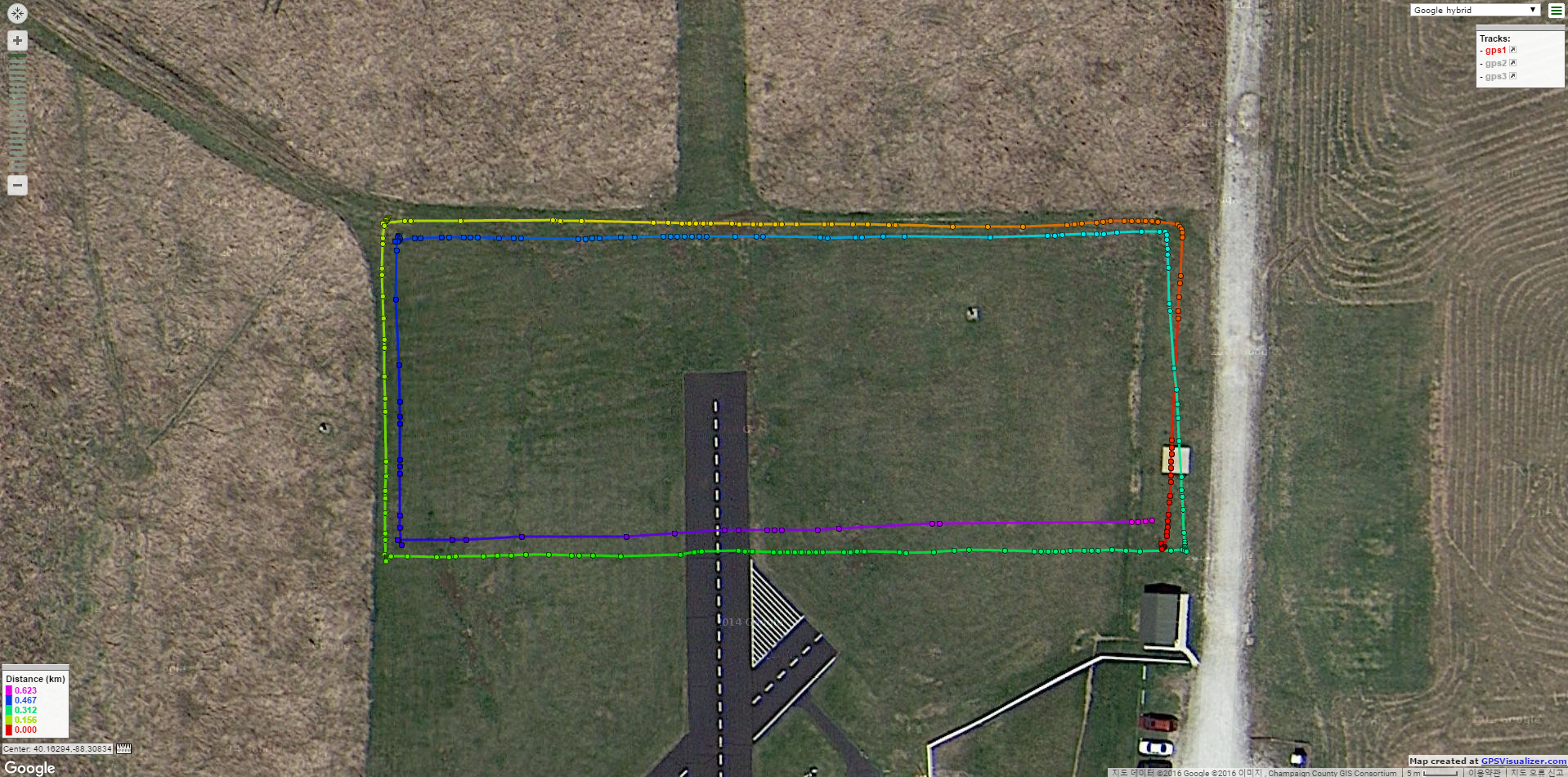}
		\caption{drone's flight path.}
		\label{fig:path1}
    \end{subfigure}
    \caption{Drone.}
    \label{fig:drone}
\end{figure}

We used daily collected groundwater data provided by the United States Geological Survey (USGS)\footnote{https://waterdata.usgs.gov}. Dataset is composed of various parameters from the US territories and we used depth to water level from the regions other than Hawaii and Alaska. Regions where the data was collected over 28 years (1987-2015) were selected and those that have unrecorded periods longer than two months were excluded. Empty records shorter than two months were filled by interpolation. Final dataset contains records from 88 sites of 10,228 days.

%\subsection{Real-world Drone Data}\label{ssec:drone} %data sample 125Hz. not strictly regular.
\subsubsection{Drone Data}
\label{sssec:drone}
We used a quadcopter%, shown in Figure~\ref{drone}, 
 as our experimental platform to collect flight sensor data. Quadcopters are aerodynamically unstable and their actuators, i.e., the motors, must be controlled directly by an on-board computer for stable flight. We used the Pixhawk\footnote{https://pixhawk.org/} as the autopilot hardware for our quadcopter. It has on-board sensors such as inertial measurement unit (IMU), compass, and barometer. We run the open-source PX4 autopilot software suite\footnote{http://px4.io/} on the ARM Cortex M4F processor on the Pixhawk. It combines sensor data and flight commands to compute correct outputs to the motors, which then controls the vehicle's orientation and position. 

We collected flight data of the quadcopter using PX4's autopilot logging facility. Each flight data is composed of time-stamped sensor and actuator measurements, flight set points (attitude, position), and other auxiliary information (radio input, battery status, etc.). We collected data sets by flying the quadcopter in an autonomous mode, in which it flies along a pre-defined path. We obtained three sets of logs by varying the path as shown in Figure~\ref{fig:drone}. 
%\manki{This part may change depending on the experiment setup; used all or not. what is the training set. How many fields are used? What are excluded?  }
In total, we used 148 sensors of 12,654 time points  excluding those that do not show any change during the flight and have missing values.

% \subsubsection{Stock Market Data}\label{sssec:stock}
\subsection{Results}\label{ssec:experimentResults}
We built group CNN and group RCNN using both spectral clustering method (explicit) and the clustering coefficient method (coeff), and compared the performance with corresponding vanilla CNNs and vanilla RCNNs. The deep CNN model architecture  are shown in Table~\ref{table:architecture}. All the learning parameters such as the learning rate and weight initialization parameters are matched. Every models were trained for 200 epochs and the best results were chosen. 

\begin{figure}[h!]
	\begin{subfigure}{\linewidth}
    	\centering
		\includegraphics[width=\linewidth]{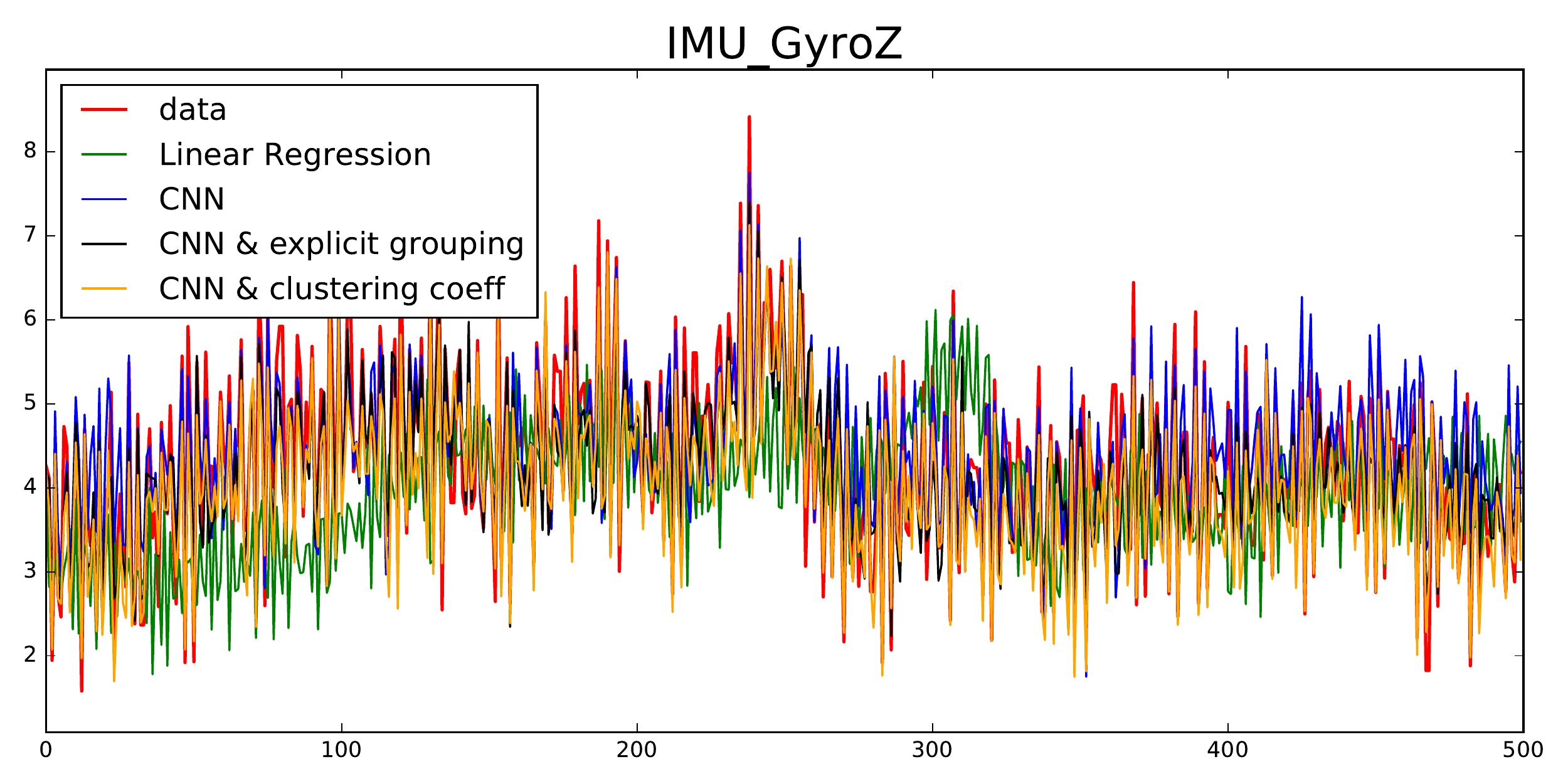}
		\caption{Drone IMU\_GyroZ sensor.}
		%\label{syn1_test}
    \end{subfigure}
    \begin{subfigure}{\linewidth}
    	\centering
		\includegraphics[width=\linewidth]{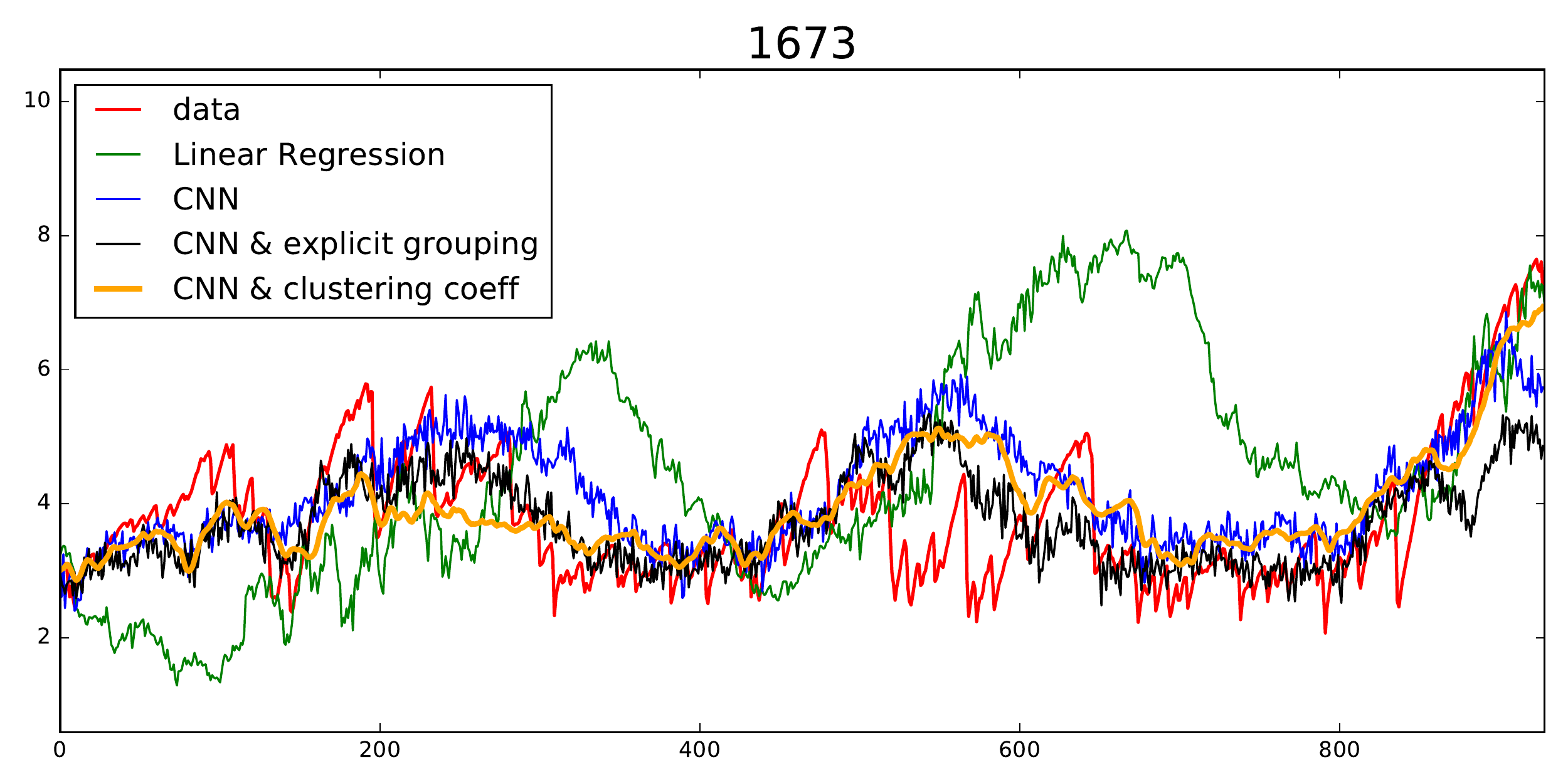}
		\caption{Groundwater site 1673.}
		%\label{syn1_test}
    \end{subfigure}
    \caption{Reconstruction examples on the test data.}
    \label{fig:testoutput}
\end{figure}

Experiment results are shown in the following tables, Table~\ref{table:experimentResult_water} and Table~\ref{table:experimentResult_drone}. In general, our group CNN models outperform in the groundwater dataset. RCNN with clustering coefficient model performs best with 0.754 SRMSE compared to 0.985 of vanilla RCNN model. Our group CNN models also tend to perform better than the vanilla CNN models in the drone flight dataset. RCNN with spectral clustering model performs best with 0.438 SRMSE compared to  0.464 SRMSE of vanilla CNN model. The values predicted by our models are shown in Figure~\ref{fig:testoutput}.

\begin{table}[h!]
\centering
  \begin{tabular}{ |l||r|r|  }
     \hline
     %\multicolumn{3}{|c|}{Country List} \\
     %\hline
     SRMSE & Mean & STDEV\\
     \hline
     \hline
     Linear Regression 				  & $1.298$ & $0.469$ \\
     Ridge Regression			  	  & $1.298$ & $0.469$ \\
     \hline
     CNN					 				& $0.999$ & $0.340$\\
     RCNN\textsuperscript{1} 	   & $0.985$ & $0.408$\\
     \hline
     CNN \& explicit 	 				& $0.861$ & $0.195$\\
     RCNN\textsuperscript{1} \& explicit & $0.929$ & $0.347$\\
     CNN \& coeff 				 		& $0.783$ & $0.254$\\
     RCNN \& coeff 				 		& $\bf{0.754}$ & $0.217$\\
     \hline
  \end{tabular}\\
  \raggedbottom{\textsuperscript{1}\footnotesize{RCNN with three RCLs of iteration 2-2-2.}}
  \caption{SRMSE of regression on Groundwater data.}
  \label{table:experimentResult_water}
\end{table}

\begin{table}[h!]
\centering
  \begin{tabular}{ |l||r|r|  }
     \hline
     %\multicolumn{3}{|c|}{Country List} \\
     %\hline
     SRMSE & Mean & STDEV\\
     \hline
     \hline
     Linear Regression 				  & $0.690$ & $0.265 $\\
     Ridge Regression			  	  & $0.731$ & $0.307$\\
     \hline
     CNN					 				& $0.464$ & $0.111$\\
     RCNN\textsuperscript{1} 	   & $0.465$ & $0.090$\\
     \hline
     CNN \& explicit 	 				& $0.479$ & $0.098$\\
     RCNN\textsuperscript{1} \& explicit & ${\bf 0.438}$ & $0.103$\\
     CNN \& coeff 				 		& $0.460$ & $0.119$\\
     RCNN \& coeff 				 		& $0.499$ & $0.061$\\
     \hline
  \end{tabular}\\
  \raggedbottom{\textsuperscript{1}\footnotesize{RCNN with three RCLs of iteration 2-2-2.}}
  \caption{SRMSE of regression on Drone data.}
  \label{table:experimentResult_drone}
\end{table}

\section{Conclusion}
In this paper, we presented two structure learning algorithms for deep CNN models. Our algorithms exploited the covariance structure over multiple time series to partition input volume into groups. The first algorithm learned the group CNN structures explicitly by clustering individual input sequences. The second algorithm learned the group CNN structures implicitly from the error backpropagation. 
In the experiments with two real-world datasets, we demonstrate that our group CNN models outperformed the existing CNN based regression methods.

% % Acknowledgements should only appear in the accepted version. 
% %\section*{Acknowledgements} 
% % \textbf{Do not} include acknowledgements in the initial version of the paper submitted for blind review. If a paper is accepted, the final camera-ready version can (and probably should) include acknowledgements. In this case, please place such acknowledgements in an unnumbered section at the end of the paper. Typically, this will include thanks to reviewers who gave useful comments, to colleagues who contributed to the ideas,  and to funding agencies and corporate sponsors that provided financial support.  
\section*{Acknowledgements}
The authors would like to thank Bo Liu at the Intelligent Robotics Laboratory, University of Illinois, for helping with collecting drone sensor data and the anonymous reviewers for their helpful and constructive comments. 
This work was supported by the National Research Foundation of Korea (NRF) grant funded by the Korea government (Ministry of Science, ICT \& Future Planning, MSIP) (No. 2014M2A8A2074096).%, and the POSCO grant (No. 2016X043).

% In the unusual situation where you want a paper to appear in the
% references without citing it in the main text, use \nocite
\nocite{Masci11}
%\clearpage
\bibliography{GRCNN}

\begin{thebibliography}{37}
\providecommand{\natexlab}[1]{#1}
\providecommand{\url}[1]{\texttt{#1}}
\expandafter\ifx\csname urlstyle\endcsname\relax
  \providecommand{\doi}[1]{doi: #1}\else
  \providecommand{\doi}{doi: \begingroup \urlstyle{rm}\Url}\fi

\bibitem[Abdel-Hamid et~al.(2012)Abdel-Hamid, Mohamed, Jiang, and
  Penn]{CNNspeechrecognition}
Abdel-Hamid, O., Mohamed, A., Jiang, H., and Penn, G.
\newblock Applying convolutional neural networks concepts to hybrid nn-hmm
  model for speech recognition.
\newblock In \emph{Proceedings of IEEE international conference on Acoustics,
  speech and signal processing}, pp.\  4277--4280, 2012.

\bibitem[Bach \& Jordan(2004)Bach and Jordan]{LearningSpectralClustering}
Bach, F.~R. and Jordan, M.~I.
\newblock Learning spectral clustering.
\newblock In \emph{Proceedings of Advances in neural information processing
  systems}, volume~16, pp.\  305--312, 2004.

\bibitem[Bengio et~al.(1994)Bengio, Simard, and Frasconi]{Bengio94}
Bengio, Y., Simard, P., and Frasconi, P.
\newblock Learning long-term dependencies with gradient descent is difficult.
\newblock In \emph{Proceedings of IEEE transactions on neural networks}, pp.\
  157--166, 1994.

\bibitem[Bourlard \& Kamp(1988)Bourlard and Kamp]{AutoBourlard1988}
Bourlard, H. and Kamp, Y.
\newblock Auto-association by multilayer perceptrons and singular value
  decomposition.
\newblock \emph{Biological cybernetics}, 59\penalty0 (4):\penalty0 291--294,
  1988.

\bibitem[Collobert \& Weston(2008)Collobert and Weston]{Collobert08}
Collobert, R. and Weston, J.
\newblock A unified architecture for natural language processing: Deep neural
  networks with multitask learning.
\newblock In \emph{Proceeding of the International Conference on Machine
  learning}, pp.\  160--167. ACM, 2008.

\bibitem[Coulibaly \& Baldwin(2005)Coulibaly and
  Baldwin]{RNNhydrologicalforecasting}
Coulibaly, P. and Baldwin, C.~K.
\newblock Nonstationary hydrological time series forecasting using nonlinear
  dynamic methods.
\newblock \emph{Hydrology}, 307:\penalty0 164--174, 2005.

\bibitem[Funahashi \& Nakamura(1993)Funahashi and
  Nakamura]{RNN:dynamical-system}
Funahashi, K.~I. and Nakamura, Y.
\newblock Approximation of dynamical systems by continuous time recurrent
  neural networks.
\newblock \emph{Neural networks}, 6\penalty0 (6):\penalty0 801--806, 1993.

\bibitem[Glorot \& Bengio(2010)Glorot and Bengio]{Glorot10}
Glorot, X. and Bengio, Y.
\newblock Understanding the difficulty of training deep feedforward neural
  networks.
\newblock \emph{Aistats}, 9:\penalty0 249--256, 2010.

\bibitem[He et~al.(2015)He, Zhang, Ren, and Sun]{He15}
He, K., Zhang, X., Ren, S., and Sun, J.
\newblock Deep residual learning for image recognition.
\newblock Technical report, arXiv:1512.03385, 2015.

\bibitem[Hochreiter \& Schmidhuber(1997)Hochreiter and Schmidhuber]{RNN:LSTM}
Hochreiter, S. and Schmidhuber, J.
\newblock Long short-term memory.
\newblock 9\penalty0 (8):\penalty0 1735--1780, 1997.

\bibitem[Intrator \& Intrator(2001)Intrator and Intrator]{Intrator01}
Intrator, O. and Intrator, N.
\newblock Interpreting neural-network results: a simulation study.
\newblock \emph{Computational statistics \& data analysis}, 37\penalty0
  (3):\penalty0 373--393, 2001.

\bibitem[Kalchbrenner et~al.(2014)Kalchbrenner, Grefenstette, and
  Blunsom]{Kalchbrenner14}
Kalchbrenner, N., Grefenstette, E., and Blunsom, P.
\newblock A convolutional neural network for modelling sentences.
\newblock \emph{arXiv:1404.2188}, 2014.

\bibitem[Koutnik et~al.(2014)Koutnik, Greff, Gomez, and
  Schmidhuber]{ClockworkRNN}
Koutnik, J., Greff, K., Gomez, F., and Schmidhuber, J.
\newblock A clockwork rnn.
\newblock \emph{arXiv preprint arXiv:1402.3511}, 2014.

\bibitem[Krause et~al.(2008)Krause, Singh, and
  Guestrin]{SensorPlacementGPKrause2008}
Krause, A., Singh, A., and Guestrin, C.
\newblock Near-optimal sensor placements in gaussian processes: Theory,
  efficient algorithms and empirical studies.
\newblock \emph{Machine Learning Research}, 9\penalty0 (Feb):\penalty0
  235--284, 2008.

\bibitem[Krizhevsky et~al.(2012)Krizhevsky, Sutskever, and
  Hinton]{Krizhevsky12}
Krizhevsky, A., Sutskever, I., and Hinton, G.~E.
\newblock Imagenet classification with deep convolutional neural networks.
\newblock In \emph{Proceedings of Advances in Neural Information Processing
  Systems}, pp.\  1097--1105, 2012.

\bibitem[Lai et~al.(2015)Lai, Xu, Liu, and Zhao]{Lai15}
Lai, S., Xu, L., Liu, K., and Zhao, J.
\newblock Recurrent convolutional neural networks for text classification.
\newblock In \emph{Proceedings of the Twenty-Ninth AAAI Conference on
  Artificial Intelligence}, pp.\  2267--2273, 2015.

\bibitem[LeCun \& Bengio(1995)LeCun and Bengio]{LeCun95}
LeCun, Y. and Bengio, Y.
\newblock Convolutional networks for images, speech, and time-series.
\newblock \emph{The handbook of brain theory and neural networks},
  3361\penalty0 (10):\penalty0 255--258, 1995.

\bibitem[LeCun et~al.(1998)LeCun, Bottou, Bengio, and Haffner]{LeCun98}
LeCun, Y., Bottou, L., Bengio, Y., and Haffner, P.
\newblock Gradient-based learning applied to document recognition.
\newblock In \emph{Proceedings of the IEEE}, volume~86, pp.\  2278--2324, 1998.

\bibitem[Liang \& Hu(2015)Liang and Hu]{Liang15}
Liang, M. and Hu, X.
\newblock Recurrent convolutional neural network for object recognition.
\newblock In \emph{Proceedings of the IEEE Conference on Computer Vision and
  Pattern Recognition}, pp.\  3367--3375, 2015.

\bibitem[Long et~al.(2015)Long, Shelhamer, and Darrell]{Long15}
Long, J., Shelhamer, E., and Darrell, T.
\newblock Fully convolutional networks for semantic segmentation.
\newblock In \emph{Proceedings of IEEE Conference on Computer Vision and
  Pattern Recognition}, pp.\  3431--3440, 2015.

\bibitem[Luxburg(2007)]{TutorialonSpectralClustering}
Luxburg, U.~V.
\newblock A tutorial on spectral clustering.
\newblock \emph{Statistics and computing}, 17:\penalty0 395--416, 2007.

\bibitem[Malhotra et~al.(2015)Malhotra, Vig, Shroff, and
  Agarwal]{LSTManomalydetection}
Malhotra, P., Vig, L., Shroff, G., and Agarwal, P.
\newblock Long short term memory networks for anomaly detection in time series.
\newblock In \emph{Proceedings of European Symposium on Artificial Neural
  Networks}, pp.\ ~89, 2015.

\bibitem[Masci et~al.(2011)Masci, Meier, Cireşan, and Schmidhuber]{Masci11}
Masci, J., Meier, U., Cireşan, D., and Schmidhuber, J.
\newblock Stacked convolutional auto-encoders for hierarchical feature
  extraction.
\newblock In \emph{Proceedings of International Conference on Artificial Neural
  Networks}, pp.\  52--59, 2011.

\bibitem[Mo \& Sinopoli(2015)Mo and
  Sinopoli]{SecureEstimationWIthIntegrityAttacks}
Mo, Yilin and Sinopoli, Bruno.
\newblock Secure estimation in the presence of integrity attacks.
\newblock \emph{IEEE Transactions on Automatic Control}, 60\penalty0 (4), 2015.

\bibitem[Mozer(1993)]{MultiscaleTemporalStructure}
Mozer, M.~C.
\newblock Induction of multiscale temporal structure.
\newblock In \emph{Proceedings of Advances in Neural Information Processing
  Systems}, pp.\  275--275. MORGAN KAUFMANN PUBLISHERS, 1993.

\bibitem[Ordóñez \& Roggen(2016)Ordóñez and Roggen]{Ordóñez16}
Ordóñez, F.~J. and Roggen, D.
\newblock Deep convolutional and lstm recurrent neural networks for multimodal
  wearable activity recognition.
\newblock \emph{Sensors}, 16\penalty0 (1):\penalty0 115, 2016.

\bibitem[Pajic et~al.()Pajic, Weimer, Bezzo, Tabuada, Sokolsky, Lee, and
  Pappas]{AttackResilientStateEstimator}
Pajic, M., Weimer, J., Bezzo, N., Tabuada, P., Sokolsky, O., Lee, I., and
  Pappas, G.~J.
\newblock Robustness of attack-resilient state estimators.
\newblock In \emph{Proceedings of the ACM/IEEE International Conference on
  Cyber-Physical Systems}. IEEE Computer Society.

\bibitem[Pascanu et~al.(2013)Pascanu, Mikolov, and Bengio]{TrainingRNN}
Pascanu, R., Mikolov, T., and Bengio, Y.
\newblock On the difficulty of training recurrent neural networks.
\newblock In \emph{Proceedings of the 30 th International Conference on Machine
  Learning}, volume~28, pp.\  1310--1318, 2013.

\bibitem[Pinheiro \& Collobert(2014)Pinheiro and Collobert]{Pinheiro14}
Pinheiro, P. and Collobert, R.
\newblock Recurrent convolutional neural networks for scene labeling.
\newblock In \emph{Proceedings of the 31 st International Conference on Machine
  Learning}, pp.\  82--90, 2014.

\bibitem[Ren et~al.(2015)Ren, He, Girshick, and Sun]{Ren15}
Ren, S., He, K., Girshick, R., and Sun, J.
\newblock Faster r-cnn: Towards real-time object detection with region proposal
  networks.
\newblock In \emph{Proceedings of Advances in neural information processing
  systems}, pp.\  91--99, 2015.

\bibitem[Rumelhart et~al.()Rumelhart, Hinton, and Williams]{RNN:original}
Rumelhart, David~E, Hinton, Geoffrey~E, and Williams, Ronald~J.
\newblock Learning representations by back-propagating errors.
\newblock \emph{Cognitive modeling}, 5\penalty0 (3):\penalty0 1.

\bibitem[Shi \& Malik(2000)Shi and Malik]{NormalizedCuts}
Shi, J. and Malik, J.
\newblock Normalized cuts and image segmentation.
\newblock In \emph{Proceedings of {IEEE} Transactions on pattern analysis and
  machine intelligence}, volume~22, pp.\  888--905, 2000.

\bibitem[Wagner \& Wagner(1993)Wagner and Wagner]{MinCutandGraphBisection}
Wagner, D. and Wagner, F.
\newblock Between min cut and graph bisection.
\newblock Springer, 1993.

\bibitem[Williams \& Zipser(1989)Williams and Zipser]{RNN:tutorial}
Williams, R.~J. and Zipser, D.
\newblock A learning algorithm for continually running fully recurrent neural
  networks.
\newblock 1\penalty0 (2):\penalty0 270--280, 1989.

\bibitem[Yang et~al.(2015)Yang, Nguyen, San, Li, and Krishnaswamy]{Yang15}
Yang, J.~B., Nguyen, M.~N., San, P.~P., Li, X.~L., and Krishnaswamy, Sh.
\newblock Deep convolutional neural networks on multichannel time series for
  human activity recognition.
\newblock In \emph{Proceedings of the International Joint Conference on
  Artificial Intelligence}, pp.\  25--31, 2015.

\bibitem[Zemel(1994)]{AutoencodersZemel1994}
Zemel, R.~S.
\newblock Autoencoders, minimum description length and helmholtz free energy.
\newblock Proceedings of the Neural Information Processing Systems Conference,
  1994.

\bibitem[Zheng et~al.(2014)Zheng, Liu, Chen, Ge, and
  Zhao]{CNNtimeseriesclassification}
Zheng, Y., Liu, Q., Chen, E., Ge, Y., and Zhao, J.~Leon.
\newblock Time series classification using multi-channels deep convolutional
  neural networks.
\newblock In \emph{Proceedings of International Conference on Web-Age
  Information Management}, pp.\  298--310. Springer International Publishing,
  2014.

\end{thebibliography}
\bibliographystyle{icml2017}

\end{document}